\documentclass[sigconf]{acmart}

\usepackage{enumerate}

\AtBeginDocument{%
  \providecommand\BibTeX{{%
    \normalfont B\kern-0.5em{\scshape i\kern-0.25em b}\kern-0.8em\TeX}}}

\copyrightyear{2021}
\acmYear{2021}
\setcopyright{acmcopyright}\acmConference[MMAsia '20]{ACM Multimedia Asia}{March 7--9, 2021}{Virtual Event, Singapore}
\acmBooktitle{ACM Multimedia Asia (MMAsia '20), March 7--9, 2021, Virtual Event, Singapore}
\acmPrice{15.00}
\acmDOI{10.1145/3444685.3446311}
\acmISBN{978-1-4503-8308-0/21/03}

\copyrightyear{2021}
\acmYear{2021}
\setcopyright{acmcopyright}\acmConference[MMAsia '20]{ACM Multimedia Asia}{March 7--9, 2021}{Virtual Event, Singapore}
\acmBooktitle{ACM Multimedia Asia (MMAsia '20), March 7--9, 2021, Virtual Event, Singapore}
\acmPrice{15.00}
\acmDOI{10.1145/3444685.3446311}
\acmISBN{978-1-4503-8308-0/21/03}
\begin{document}

\title{Improving auto-encoder novelty detection using channel attention and entropy minimization}

\author{Miao Tian}
\orcid{0000-0002-4926-1151}
\affiliation{
	\institution{Zhejiang University of Technology}
  	\city{Hangzhou}
  	\country{China}
  }
\email{2111812039@zjut.edu.cn}
\author{Dongyan Guo}
\authornotemark[1]
\affiliation{
  \institution{Zhejiang University of Technology}
  \city{Hangzhou}
  \country{China}}
\email{guodongyan@zjut.edu.cn}
\author{Ying Cui}
\affiliation{
  \institution{Zhejiang University of Technology}
  \city{Hangzhou}
  \country{China}
}
\email{cuiying@zjut.edu.cn}
\author{Xiang Pan}
\affiliation{
 \institution{Zhejiang University of Technology}
 \city{Hangzhou}
 \country{China}}
\email{panx@zjut.edu.cn}
\author{Shengyong Chen}
\affiliation{
  \institution{Tianjin University of Technology}
  \city{Tianjin}
  \country{China}}
\email{sy@ieee.org}

%
%


\begin{abstract}
Novelty detection is a important research area which mainly solves the classification problem of inliers which usually consists of normal samples and outliers composed of abnormal samples. Auto-encoder is often used for novelty detection. However, the generalization ability of the auto-encoder may cause the undesirable reconstruction of abnormal elements and reduce the identification ability of the model. To solve the problem, we focus on the perspective of better reconstructing the normal samples as well as retaining the unique information of normal samples to improve the performance of auto-encoder for novelty detection. Firstly, we introduce attention mechanism into the task. Under the action of attention mechanism, auto-encoder can pay more attention to the representation of inlier samples through adversarial training. Secondly, we apply the information entropy into the latent layer to make it sparse and constrain the expression of diversity. Experimental results on three public datasets show that the proposed method achieves comparable performance compared with previous popular approaches.
\end{abstract}

%



\maketitle

\section{Introduction}
\begin{figure}
	\centering
	\includegraphics[scale=.4]{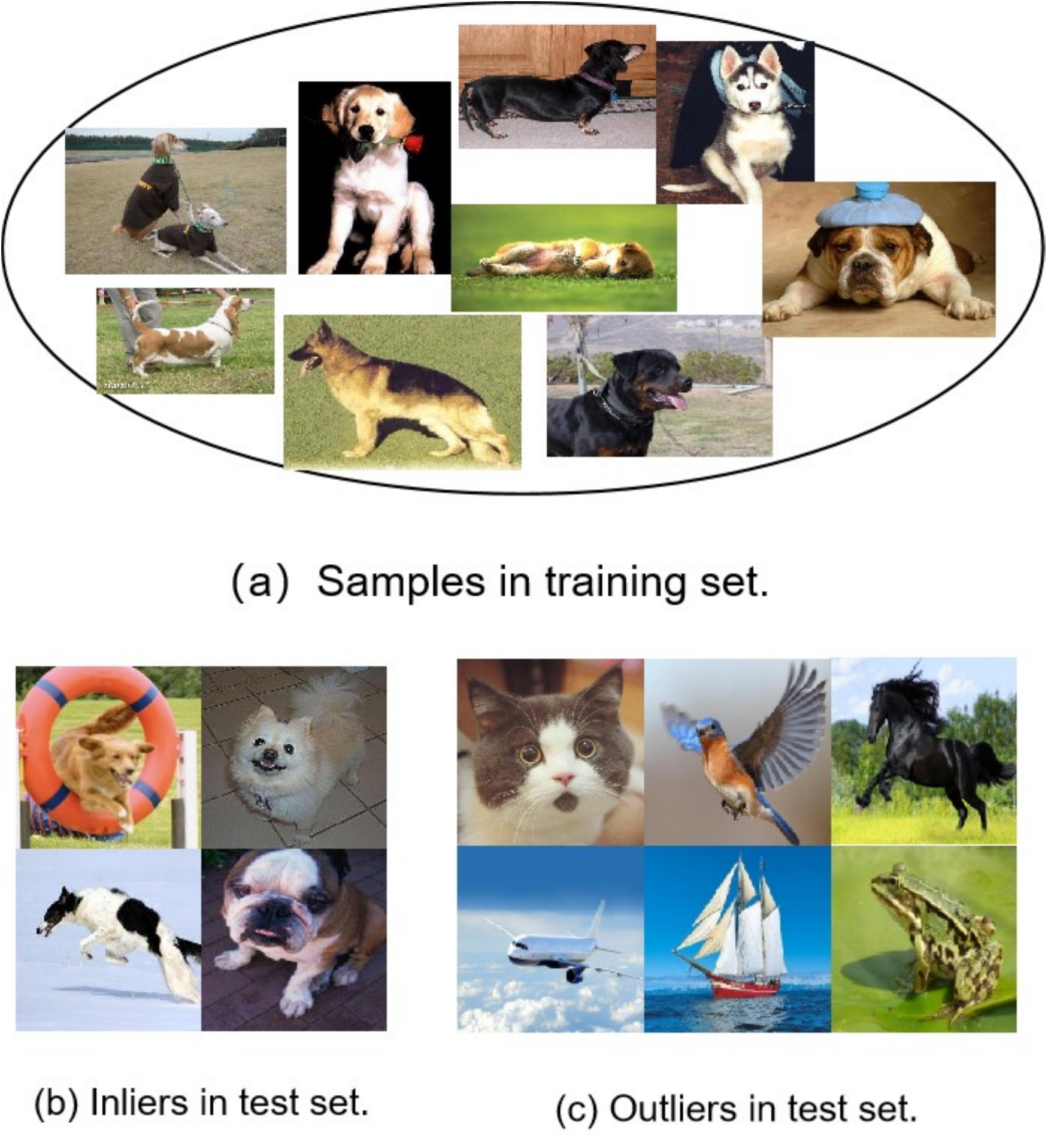}
	\caption{One-class novelty detection.The model is trained on inlier samples (a) to detect inliers and outliers. Figure (a) shows some inlier examples which are included in the training dataset. Figure (b) and Figure (c) show some inlier examples and outlier examples in the test dataset respectively.}
	\label{pic1}	 
\end{figure}

%
%
Novelty detection is an important and challenging task in computer vision. It mainly addresses the problem of quantifying the probability of a test sample belonging to a distribution defined by a training dataset. Different from other machine learning tasks, in novelty detection only a single class of samples can be observed during training. Besides, the scarcity, variability and unpredictability of outlier samples make it difficult to make a decision, as shown in Figure \ref{pic1}.

It has many applications on anomaly detection \cite{Anogan,Efficentgan,an2015variational,akccay2019skip}, intruder detection \cite{oza2019active,perera2018dual,perera2019learning}, and biomedical data processing \cite{roberts1999novelty}. Especially in some security critical areas, novelty detection is often utilized as an important preprocessing step to detect abnormal patterns in advance. The key of novelty detection is to make the model have a better representation of inlier samples and a degree of \textit{surprisal} \cite{tribus1961thermostatics} to outlier samples after only learning from inlier samples. Generally speaking, both of these abilities should be strengthened in training. A assumption is that outlier samples differ from inlier samples not only in high dimensional data space but also in low dimensional latent space \cite{akcay2018ganomaly,akccay2019skip}. Therefore, the significance of this problem is how to obtain a better reconstruction of high dimensional data space and low dimensional latent space. In view of this, many researchers adopt the auto-encoder to jointly learn the generation of a high dimensional image space and the low dimensional latent space for anomaly detection. The detection model is comprised of generator and discriminator. However, existing methods regard multi-class targets as normal in problem setting but neglect the data distribution of latent space. 

The basic principle of novelty detection is that the model has a high ability to reconstruct the normal test samples but restrain the reconstruction of abnormal samples. Consequently, the reconstruction error can be used as the standard to calculate the novelty score of the test input. However, the generalization ability of the auto-encoder may lead to the success reconstruction of abnormal elements and reduce the identification ability of the model \cite{OCGAN}.
 
In view of the characteristics of the task, we focus on how to better reconstruct the normal samples and retain the unique information of normal samples. In order to enhance the expression of normal samples and improve the detection effect, we introduce the channel attention mechanism into network structure to mine the interested features and realize adaptive feature refinement. Secondly, from the perspective of information theory, we introduce the information entropy to the latent layer. The expression of diversity can be constrained by reducing the information entropy of the latent layer, while the unique information of normal samples can be retained. Experimental results on three public datasets demonstrate the effectiveness of the proposed method. The main contributions of this paper are as follows:

\begin{enumerate}[1)]
	\item {\bfseries Features selection:} The channel attention mechanism is applied to make the network focus on the important and interesting features.
	\item {\bfseries Latent space Constraining:} The information entropy is introduced into the latent layer to make it sparse and constrain the expression of diversity.
\end{enumerate}



\section{Related Work}
Novelty detection and anomaly detection are highly correlated. Due to the scarcity and unpredictability of anomaly samples, unsupervised learning is generally used to solve the problem. Some traditional solutions include one-class SVM \cite{OCSVM} and LOF \cite{LOF}. However, the traditional methods depend on the distribution of normal and abnormal samples in training. Nowadays, with the great achievements of deep learning in the field of image processing, the main strategies are based on the reconstruction method and probability distribution method by introducing deep neural networks to this problem. 

In the reconstruction-based strategy, the reconstruction error is used to measure the novelty of samples. AnoGAN \cite{Anogan} first apply GAN \cite{GAN} into image anomaly detection. The best potential space \textit{z} is found by back-propagation method, and the anomaly is judged according to the reconstructed image and the original image.  Besides, the auto-encoder is also exploited to model and measure the reconstruction error. Model testing is time-consuming. In GANomaly \cite{akcay2018ganomaly}, the model adopts the structure of generator and discriminator. The generator is encoder-decoder-encoder sub-network which enables the model to map the input to a lower dimension vector, which is then used to reconstruct the generated output image. The use of the additional encoder network maps this generated image to its latent representation. Minimizing the distance between these images and the latent vectors during training aids in learning the data distribution for the normal samples. In Skip-GANomaly \cite{akccay2019skip}, as an upgrade of the previous method, the generator of model adopts UNet \cite{Unet} style network for better image reconstruction and the previous additional encoder network is removed. The penultimate layer of the discriminator is used as a feature extractor to output low dimensional feature information. Besides, the discriminator is also used to identify the generated image from the real image. They learn from each other in the form of adversarial training. In ALOCC \cite{ALOCC}, the model consists of a generator and a discriminator. The auto-encoder is utilized as the generator and plays two roles, one is to enhance normal samples, and the other is to disturb abnormal samples. Furtherly, in \cite{future}, they propose for the first time to use the predicted future frame to compare with the current frame for anomaly detection. It adds temporal constraint in the video anomaly detection. The generator of model adopts UNet \cite{Unet} structure to predict next frame.

In the probability distribution-based strategy, more attention is paid to the distribution of low-dimensional latent layer vectors. For instance, in \cite{latent}, it equips a deep autoencoder with a parametric density estimator to learns the probability distribution of potential representation through the autoregressive process. In OCGAN \cite{OCGAN}, the denoising auto-encoder network is utilized as the generator. By using the adversarial training mechanism, the inlier samples and low-dimensional latent space can achieve one-to-one mapping to exclusively represent given inliers. Recently, self-supervised learning has been introduced to novelty detection. The main idea is taking advantage of some auxiliary tasks to better describe the data distribution of normal samples. In \cite{geom}, dozens of geometric transformations such as rotation, fold and translation, are applied on samples to train multi-class model and this auxiliary task is used to obtain the feature detector to identify anomalies. In \cite{Hendrycks2019Using}, the normal samples are rotated in four angles of 0, 90,180,270, one label corresponds to one rotation type to build an auxiliary self-supervised rotation loss classifier. In this paper, we consider to form a effective method based on the reconstruction-based strategy for one-class novelty detection. 

\section{Proposed method}
\begin{figure*}[h]
	\centering
	\includegraphics[scale=0.3]{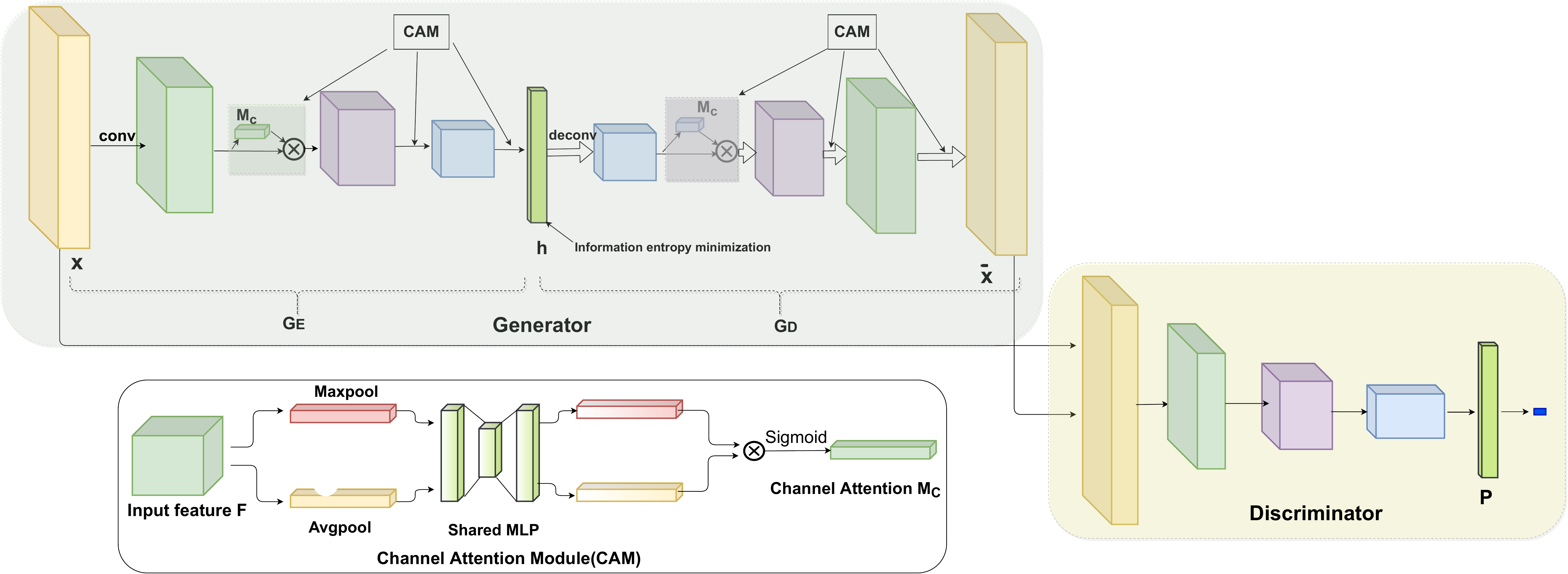}
	\caption{The proposed novelty detection framework. The overall network consists of two sub-networks: a generator G and a discriminator D. Channel attention module (CAM) is applied to the generator G to choose important features information. The latent layer $h$ of generator is constrained by information entropy minimization.}
	\label{pic2}
\end{figure*}

To define the problem, a given dataset  is divided into training set $ D_{train} $ , validation set $ D_{val} $ and test set $ D_{test} $. The data in training set $ D_{train} $ is composed of $ m $ data samples with  $ D_{train}  =  \left\{(x_1,y_1),(x_2,y_2)...(x_m,y_m)\right \}$, in which $ y_i=0 $ labels normal samples.
The validation set accounts for 15\% of the training set. The test set $ D_{test} = \left \{(x_1,y_1),(x_2,y_2)...(x_n,y_n)\right \}$consists of $ n $ samples, including normal samples which are labeled by $ y_i=0 $ and abnormal samples which are labeled by $ y_i=1 $.

In Skip-GANomaly \cite{akccay2019skip}, the goal is to acquire the high-dimensional and low-dimensional image features. It is trained by the means of unsupervised adversarial learning. The following three loss functions are adopted to train the model. An adversarial loss $L_{adv} $ is used to constrain the generator $ G $ and discriminator $ D $ so that they can learn from each other:
\begin{equation}
L_{adv}=\mathop{\mathbb{E}}\limits_{x\sim p_{x}}[logD(x)]+\mathop{\mathbb{E}}\limits_{x\sim p_{x}}[log(1-D(G(x))]
\end{equation}
where $ p_x $ represents the data distribution of training set.
Then, in order to make the generated image $ \hat{x} $ similar to the original image $ x $, the context information is obtained by the reconstruction error denoted as a context loss $ L_{con} $:
\begin{equation}
L_{con}=\mathop{\mathbb{E}}\limits_{x\sim p_{x}}\left | x - \hat{x}\right |_{1}
\end{equation}
Similarly, low dimensional features of the generated image and the original image are expected to be as close as possible by a feature match loss $ L_{fea} $:
\begin{equation}
L_{fea}=\mathop{\mathbb{E}}\limits_{x\sim p_{x}}\left | f(x)-f(\hat{x})\right|_{2}
\end{equation}
Note that the feature $ f(\cdot ) $ here refers to the penultimate output in the discriminator which is marked with \textit{P} in the Figure \ref{pic2}.

During the test, it is observed that the ability of network feature extraction is very important. However, existing  methods like Skip-GANomaly \cite{akccay2019skip} neglect to pay attention to the expression of latent space. In novelty detection setting, it only focuses on the representation of one class of samples. Desired goal is other class samples have a poor reconstruction after the generator. Therefore, it is particularly important for the model to only focus on the expression of normal samples. Considering the selection of important features and the constrain of latent space , we propose to introduce channel attention mechanism and information entropy minimization into the network based on the usage of de-noising auto-encoder to achieve certain constraints on auto-encoder.

The proposed novelty detection framework is mainly composed of two parts: a generator \textit{G} and a discriminator \textit{D}, as shown in Figure \ref{pic2}. The generator \textit{G} is a de-noising auto-encoder, which is responsible for the generation of images. The discriminator \textit{D} is a plain identification network, which is responsible for the identification of images. The generator and the discriminator are trained against to learn from each other. In the generator, the network is a symmetrical denosing auto-encoder including encoder $ G_E $ and decoder $ G_D $. When an image added with random noise is put into the network, a latent layer $ \textit{h} $ is obtained. Then, the information entropy of latent layer $ \textit{h} $ is calculated. Correspondingly, the generated image $ \hat{x} $ is obtained by decoder $ G_D $. The network structure of encoder are composed of stacked convolution blocks including convolution, batchnorm and activation function. Channel attention mechanism is added after each block.  The decoder are composed of a series of deconvolution, batchnorm and activation function in each convolution blocks and similarly channel attention operations is added. The network structure of discriminator is similar with encoder of DCGAN \cite{dcgan}. Besides, the latent layer $ \textit{h} $ of generator is constrained by information entropy minimization. 

{\bfseries Information entropy.} In the novelty detection, there is only one kind of training samples. When the samples passing through the encoder $ G_E $ of the generator, the compressed information will be saved in the latent layer $ \textit{h} $, and then the generated image will be obtained through the decoder $ G_D$. The information of latent layer $ \textit{h} $ is generally considered to retain the feature information of the training class. Since during the training process, only normal samples can be observed in the model, the latent layer is hoped to show a lower entropy in the possible representation so as to extract features that are easy to predict and repeat in inliers. From the constrain of latent space and preserving unique information, we optimize this layer by minimizing the information entropy, which can be utilized to measure the amount of information and describe the uncertainty of information source. Before calculating the information entropy of the latent layer \textit{h}, the sum of the latent layer is changed to 1 via Softmax activation function. The length of latent layer $ \textit{h} $ is denoted as $ t $. Finally, the information entropy loss function is defined as:
\begin{equation}
L_{inf}={\sum_{i=1}^t}-h_i\cdot \log (h_i)
\end{equation}

{\bfseries Attention mechanism.} 
In order to increase the representation ability of the network for paying more attention to the important and interesting features, we consider introducing the attention mechanism to the network. As a complement to convolution operation, the attention mechanism has shown great ability in natural language and image processing. In recent years, there are many works show that the attention mechanism plays an important role in refining feature selection. In \cite{Residual}, the classification ability of the network is improved by introducing the attention mechanism to obtain the refined feature map. In SAGAN \cite{SAGAN}, in order to make the generated image more realistic, it assigns weights to each element of the middle feature map to focus on the global dependency. Moreover, in CBAM \cite{CBAM}, there are two forms of attention mechanism, one is channel attention mechanism, which refines the feature map by assigning weights to each channel of the feature map, and the other is spatial attention mechanism, which pays more attention to the spatial relationship between features.  In novelty detection, it pays attention to the content information instead of the location of the target and specific pixel information. 
We take notice of the global semantic information, which can obtain by the global maximum pooling and average pooling operators. Therefore we choose to introduce the channel attention mechanism \cite{CBAM} into the network for effective feature selection.

As shown in Figure \ref{pic2}, we add the channel attention mechanism to the generator \textit{G} to choose important features information. Specifically, given a feature map \textit{F}, two vectors $ V_1 $ and $ V_2 $ are obtained through the max-pooling and average-pooling operations respectively. After that, the vectors pass through a shared three-layer fully connected network denoted as MLP in turn. The dimensions of the output layer and the input layer are the same, and the number of neurons in the middle hidden layer is controlled by the proportion \textit{R}. Finally, the corresponding elements of the two outputs are added and the weight $ M_C $ of each channel is obtained by the activation function Sigmoid denoted as $ \zeta $ . The given feature map \textit{F} is multiplied by the corresponding channel weight $ M_C $ to get the refined feature map ${F}'$ , which can be formulated as follows:

\begin{equation}
{F}'=\zeta(MLP(AvgPool(F))+MLP(MaxPool(F)))\cdot F
\end{equation}

Finally, the model is optimized to minimize the  integrated loss function \textit{Loss}: 

\begin{equation}
Loss=\lambda_{adv} L_{adv}+\lambda_{con} L_{con}+\lambda_{fea} L_{fea}+\lambda_{inf} L_{inf}
\end{equation}
where $ L_{adv} $, $ L_{con} $, $ L_{fea} $, $ L_{inf} $ are the weighting parameters which are used to adjust the impact of individual loss to the overall objective function. 

\section{Experimental Results}

\subsection{Datasets and Experimental Results}
\begin{figure}[h]
	\centering
	\includegraphics[scale=.42]{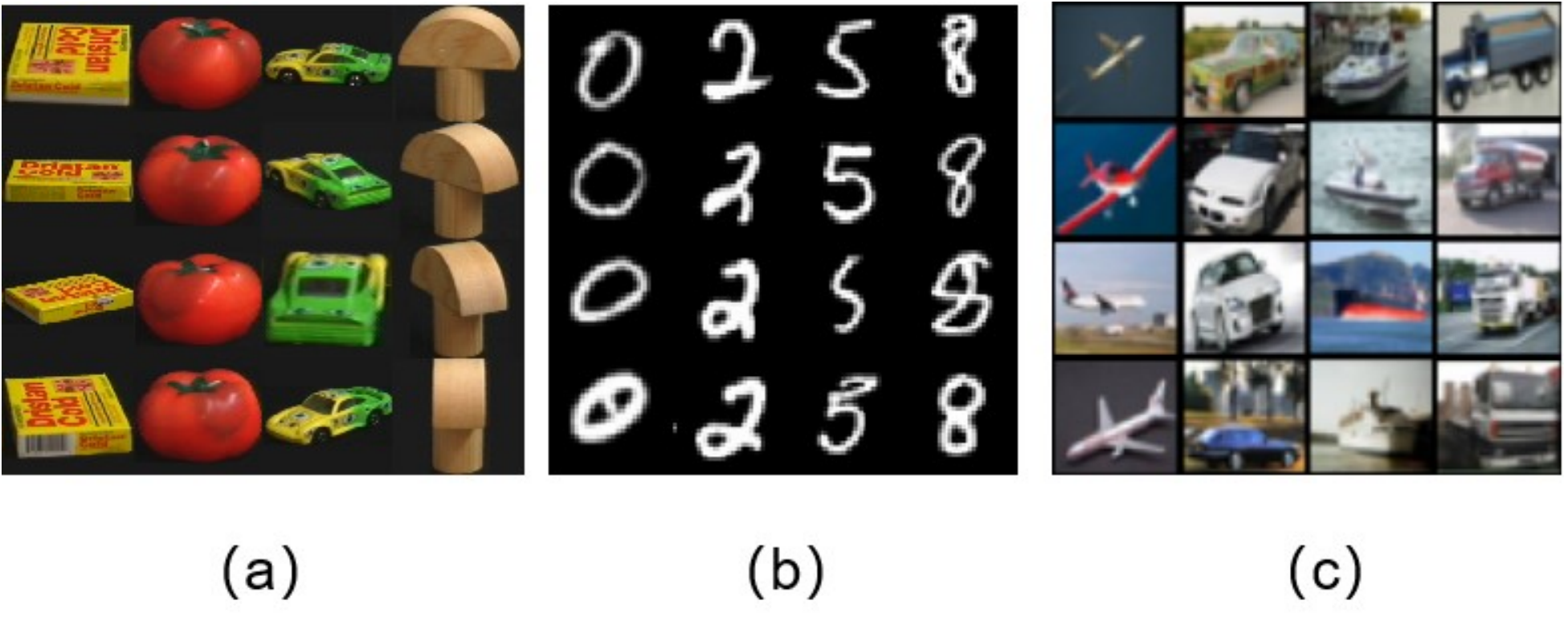}
	\caption{Typical images from the datasets used for evaluation. Each column represents a class. Figure (a) belongs to COIL-100 dataset. Figure (b) is classical MNIST dataset. Figure (c) is natural image dataset CIFAR10.}
	\label{pic3}
\end{figure}

We use Adam optimizer for model training. The learning rate is set to 1e-3. We train the model until convergence of loss function with a batch size of 64 for MNIST and CIFAR10 datasets  and a batch size of 15 for COIL-100 dataset.

In the inference stage, we adopt the novelty score, proposed in \cite{Anogan} and
also employed in \cite{Efficentgan}. The high-dimension context loss $ L_{con} $ and low-dimension feature loss $ L_{fea} $ are used to get the novelty score, therefore the novelty score is defined as \textit{A(x)} for a test sample x:
\begin{equation}
A(x)=\lambda L_{con}+(1-\lambda) L_{fea}
\end{equation}
By normalizing the novelty score to [0, 1], we get the novelty score $ A(x_{i}) $ of each test image. When the score value is closer to 1, it indicates that the image is more likely to be an outlier image. In this way, all novelty score values of the test set are compared with the real label to obtain the result:
\begin{equation}
	{A(x_{i})}' = \frac{A(x_{i})-min{A(x)}}{max(A(x)) - min(A(x))}
\end{equation}

In order to evaluate our approach, we conduct expensive experiments on three datasets, namely COIL-100, MNIST and CIFAR10. Figure \ref{pic3} shows some typical images of each datasets.  During all experiments, the pixel values of all images are scaled in [-1, 1]. Since the problem domain settings of Skip-ganomaly \cite{akccay2019skip} for anomaly detection and novelty detection are different, these two tasks are not comparable. Therefore, we compare our method with some traditional methods including OCSVM \cite{OCSVM}, KDE \cite{KDE} and DAE \cite{DAE}, and some deep learning methods including VAE \cite{VAE}, PixCNN \cite{van2016conditional} and AnoGAN \cite{Anogan}. The AUC (Area Under Curve) criterion is used as the measurement for performance evaluation. 
\begin{table}[h]
	\caption{The results for COIL-100 dataset}
	\begin{tabular}{lc}
		
		\toprule
		Method  & AUC   \\ 
		\midrule
		ALOCC DR \cite{ALOCC} & 0.809 \\ 
		ALOCC D \cite{ALOCC}  & 0.686 \\ 
		DCAE \cite{DCAE}     & 0.949 \\ 
		Our method    		 &0.970  \\
		\bottomrule 
	\end{tabular}
	\label{table1}
\end{table}
\begin{table*}[h]
	\caption{The results for MNIST dataset.}
	\begin{tabular}{lccccccccccc}
		\toprule
		Method & 0     & 1     & 2     & 3     & 4     & 5     & 6     & 7     & 8     & 9     & Mean  \\
		\midrule
		OCSVM \cite{OCSVM} & 0.988 & \textbf{0.999} & 0.902 & 0.950 & 0.955 & 0.968 & 0.978 & 0.965 & 0.853 & 0.955 & 0.9513 \\
		KDE \cite{KDE}  & 0.885 & 0.996 & 0.710 & 0.693 & 0.844 & 0.776 & 0.861 & 0.884 & 0.669 & 0.825 & 0.8143 \\
		DAE \cite{DAE}  & 0.894 & 0.999 & 0.792 & 0.851 & 0.888 & 0.819 & 0.944 & 0.922 & 0.740 & 0.917 & 0.8766 \\
		VAE \cite{VAE}  & \textbf{0.997} & 0.999 & 0.936 & \textbf{0.959} & \textbf{0.973} & 0.964 & \textbf{0.993} & \textbf{0.976} & 0.923 & 0.976 & \textbf{0.9696} \\
		PixCNN \cite{van2016conditional}      & 0.531 & 0.995 & 0.476 & 0.517 & 0.739 & 0.542 & 0.592 & 0.789 & 0.340 & 0.662 & 0.6183 \\
		GAN \cite{Anogan}  & 0.926 & 0.995 & 0.805 & 0.818 & 0.823 & 0.803 & 0.890 & 0.898 & 0.817 & 0.887 & 0.8662 \\
		AND \cite{AND}  & 0.984 & 0.995 & \textbf{0.947} & 0.952 & 0.960 & \textbf{0.971} & 0.991 & 0.970 & \textbf{0.922} & \textbf{0.979} & 0.9671 \\
		AnoGAN \cite{Anogan}& 0.966& 0.992 & 0.850 & 0.887 & 0.894 & 0.883 & 0.947 & 0.935 & 0.849 & 0.924 & 0.9127 \\
		Our method & 0.994 & \textbf{0.999} & 0.939 & 0.952 & 0.967 & 0.965 & 0.988 & 0.972 & 0.912 & 0.976 & 0.9665 \\
		\bottomrule
	\end{tabular}
	\label{table2}
\end{table*}
\begin{table*}
	\caption{The results for CIFAR10 dataset.}
	\begin{tabular}{lccccccccccc}
		\toprule
		Method&Plane&Car&Bird&Cat&Deer&Dog&Frog&Horse&Ship&Truck&Mean \\
		\midrule
		OCSVM \cite{OCSVM} & 0.630 & 0.440 & 0.649 & 0.487 & 0.735 & 0.500 & 0.725 & 0.533 & 0.649 & 0.508 & 0.5856 \\
		KDE \cite{KDE} & 0.658 & 0.520 & 0.657 & 0.497 & 0.727 & 0.496 & 0.758 & 0.564 & 0.680 & 0.540 & 0.6097 \\
		DAE \cite{DAE}  & 0.411 & 0.478 & 0.616 & 0.562 & 0.728 & 0.513 & 0.688 & 0.497 & 0.487 & 0.378 & 0.5358 \\
		VAE \cite{VAE}  & 0.700 & 0.386 & \textbf{0.679} & 0.535 & \textbf{0.748} & 0.523 & 0.687 & 0.493 & 0.696 & 0.386 & 0.5833 \\
		PixCNN \cite{van2016conditional} & \textbf{0.788} & 0.428 & 0.617 & 0.574 & 0.511 & 0.571 & 0.422 & 0.454 & 0.715  & 0.426 & 0.5506 \\
		GAN \cite{Anogan}  & 0.708 & 0.458 & 0.664 & 0.510 & 0.722 & 0.505 & 0.707 & 0.471 & 0.713 & 0.458 & 0.5916 \\
		AND \cite{AND} & 0.717 & 0.494 & 0.662 & 0.527 & 0.736 & 0.504 & 0.726 & 0.560 & 0.680 & 0.566 & 0.6172 \\
		AnoGAN \cite{Anogan}& 0.671& \textbf{0.547} & 0.529 & 0.545 & 0.651 & 0.603 & 0.585 & \textbf{0.625} & \textbf{0.758} & \textbf{0.665} & 0.6179 \\
		Our method & 0.692 & 0.538 & 0.654 & \textbf{0.636}& 0.745 & \textbf{0.630} & \textbf{0.762} & 0.570 & 0.736 & 0.592 & \textbf{0.6555} \\
		\bottomrule
	\end{tabular}
	\label{table3}
\end{table*}

{\bfseries COIL-100:} COIL-100 dataset is a collection of color pictures, including 100 objects taken from different views. Among them, the difference between images within a class is small, while the gap between classes is relatively large. For consistency, the training set accounts for 80\% of normal samples and the remaining 20\% of normal samples are used for test. The abnormal test samples are randomly selected, thus they make up half of the test set. We randomly take one class as the normal class and use the other classes samples as anomalous data. We repeat the experiment 20 times and finally got the average result. Because the dataset is relatively simple, the method considered produces a higher AUC value illustrated in Table \ref{table1}. In ALOCC \cite{ALOCC}, the importance of reconstruction step is reflected. The AUC value of our method is 0.961 more than DCAE \cite{DCAE} $2\%$. 

{\bfseries MNIST:} It is a classical dataset which is often used as novelty detection. It consists of ten types of handwritten digits. The size of each image is 28*28 resolutions. To accommodate the input of the model, each sample is resized to 32*32 resolutions through the method of bilinear interpolation. In training, one class is considered inliers, while the remaining classes are considered outliers. In order to be consistent with the comparison methods, we use the partition of training-testing splits of the given dataset. The normal samples of training split are used for training set and the all samples of testing split are used for testing set. The results are showed in Table \ref{table2}. We consider some traditional methods, OCSVM \cite{OCSVM} and KDE \cite{KDE} etc., and some methods based on deep learning, GAN \cite{Anogan} and GANomaly \cite{akcay2018ganomaly} etc. Each row represents the experimental results of one method, expressed with AUC value. The last column is the average AUC value of each method. Compared with the methods in Table \ref{table2}, our result has proximal effects with VAE \cite{VAE} and AND \cite{AND}. We analyse that the effects of the channel attention and information entropy has a very small effect on gray image with simple background. When the background becomes complex and diverse, the newly added module can play a more effective role. This can be verified in CIFAR10 dataset. In general, our method has achieved desirable results.

{\bfseries CIFAR10:} The dataset contains 60,000 natural images, all of which are derived from the real world. They are 32*32 resolutions and are divided into 10 categories with 6000 images in each category. In the experiment, the partition of the training set and testing set is the same as MNIST dataset. One class is considered inliers while the other nine classes are considered outliers. In Table \ref{table3}, we show the results of each class as an inlier sample. In the method based on deep learning, our method shows great advantages. Because Anogan \cite{Anogan} needs to find the best representation through backpropagation, it is quite slow. Our method can complete the judgment as long as we have a forward inference. It is worth noting that when the class of cat or dog is regarded as outliers, almost all methods show common results. Similarly, the identification ability of the considered model is not very good in the similar classes car and truck.  Relative to AnoGAN \cite{Anogan} and AND \cite{AND}, our results have achieved a markable improvement on this issue improved by 4$\%$. In AND \cite{AND}, autoregressive density estimation is used, but the method we put forward is relatively simple and can achieve better results. Through the newly added part, the model can adaptively select important features and retain unique information.

\subsection{Ablation Study}
In order to evaluate the effectiveness of the proposed model, a series of ablation studies are carried out on the natural image dataset CIFAR10. Two new components are tested respectively. Firstly, we consider the structure only consists of generator and discriminator. The generator is a denosing auto-encoder. The value of AUC is 0.619 in Table \ref{table4}. Secondly, information entropy is added to the latent layer \textit{h} to optimize loss function in generator. The experimental result increased by 2\%. It is shown that limiting the diversity of latent layer expression is conducive to the extraction of a single class of unique information. Thirdly, on the basis of adding above operation, channel attention mechanism is added into generator. The performance improves about 1.3\% which shown attention mechanism is useful to novelty detection in feature adaptive selection. The effectiveness of the new addition was confirmed by the experimental analysis of each component.
\begin{table}[h]
	\caption{The results of ablation study for CIFAR10 dataset.}
	\begin{tabular}{lc}
		\toprule
		Method          & AUC \\
		\midrule
		Generator and discriminator & 0.619 \\
		With latent entropy loss    & 0.642 \\ 
		With channel attention      & 0.655 \\
		\bottomrule
	\end{tabular}
	\label{table4}
\end{table}

\section{Conclusion}
In this paper, we note that the quality of feature extraction within one class directly affects model's the familiarity with inlier samples and the sensitivity of outlier samples. We propose to introduce channel attention mechanism into the generator to better extract inlier features. At the same time, we improve the loss function from the perspective of information entropy to constrain the expression of the latent layer and remove the redundancy of coding information. Besides, unsupervised adversarial learning is used to optimize both high-dimensional data space and low-dimensional latent space. The proposed method is validated in three open datasets. In future, we will pay more attention to the expression of the model for single class samples.
 
\begin{acks}
  This work is supported in part by  the Natural Science Foundation of China (62002325, 61802348), and in part by Natural Science Foundation of Zhejiang Province (LQ18F030013, LQ18F030014).
\end{acks}

\bibliographystyle{ACM-Reference-Format}
\bibliography{ref}
%
%
%
%
%
%
%
%
\end{document}